%% file: iclr2021_conference.tex
\documentclass{article} 
\usepackage{iclr2021_conference,times}

\input{math_commands.tex}

\usepackage{wrapfig}
\usepackage{lipsum}
\usepackage{hyperref}
\usepackage{url}
\usepackage{scrextend}
\usepackage{soul}

\usepackage{amsmath,amsfonts,amssymb,amsthm}

\usepackage{mathtools}

\title{Biological connectomes as a representation for the architecture of artificial neural networks}

\author{Samuel Schmidgall \\
George Mason University \\
\texttt{sschmidg@gmu.edu} \\
\And
Catherine Schuman \\
University of Tennessee, Knoxville \\
\texttt{cschuman@utk.edu} 
\And
Maryam Parsa \\
George Mason University \\
\texttt{mparsa@gmu.edu} \\
}

\iclrfinalcopy 
\begin{document}

\maketitle

\begin{abstract}


Grand efforts in neuroscience are working toward mapping the connectomes of many new species, including the near completion of the \textit{Drosophila melanogaster}. It is important to ask whether these models could benefit artificial intelligence. In this work we ask two fundamental questions: (1) where and when biological connectomes can provide use in machine learning, (2) which design principles are necessary for extracting a good representation of the connectome. Toward this end, we translate the motor circuit of the \textit{C. Elegans} nematode into artificial neural networks at varying levels of biophysical realism and evaluate the outcome of training these networks on motor and non-motor behavioral tasks. We demonstrate that biophysical realism need not be upheld to attain the advantages of using biological circuits. We also establish that, even if the exact wiring diagram is not retained, the architectural statistics provide a valuable prior. Finally, we show that while the \textit{C. Elegans} locomotion circuit provides a powerful inductive bias on locomotion problems, its structure may hinder performance on tasks unrelated to locomotion such as visual classification problems.
 

\end{abstract}

\section*{Introduction}

Artificial neural networks (ANNs) are typically designed to have highly general architectures that work well on a wide variety of problems, depending primarily on the fine-tuning of weights across very large amounts of data to solve a given problem. This differs from nature, where much of animal behavior is encoded in the genome and decoded via highly structured brain connectivity that allows the rapid development of useful behaviors without much experience (\cite{zador2019critique}). These structured neural architectures have been fine-tuned over millions of years by evolution to bias the organism toward behavior which is useful for survival.

This process is not too unfamiliar, as decades of artificial intelligence research has produced many models of representing data that provide strong inductive biases on a variety of task domains. Convolutional neural networks (CNNs) perform particularly well on image learning problems, with even randomly initialized CNNs can provide excellent results in certain image learning settings (\cite{ulyanov2018deep}). Recurrent Neural Networks (RNNs) are particularly well-suited for variable input and variable time based problems and Long-Short Term Memory networks (LSTMs) present recurrent structure that is well-suited for handling much larger time spans than the RNNs (\cite{lipton2015critical}).


A recent grand endeavor in neuroscience has led to the mapping of the first completed connectome of a nematode species, the \textit{Caenorhabditis Elegans} (\textit{C. Elegans}). For the first time an opportunity has arisen for translating connectomes from neuroscience into the architecture of an artificial neural network. However, it is uncertain as to whether this representation could provide value to machine learning. Furthermore, it is uncertain as to how the architecture should model the synaptic and neuronal dynamics to best represent the connectome. 

A complimentary work explores designing an ANN architecture that resembles \textit{C. Elegans}-like microcircuits by combining discrete-time integrator and oscillation units (\cite{bhattasali2022neural}). However, (a) this work did not use the \textit{C. Elegans} locomotion circuit, (b) the dynamics of the presented architecture differs from the \textit{C. Elegans} locomotion circuit (\textit{e.g. the use of sinusoidal input}), and (c) did not provide understanding of how to better utilize exact connectome data. Additionally, it did not provide either improvements nor limitations to connectome-based approaches. \textbf{The goal of our paper is not to design an architecture that works well on arbitrary problems, but to discern where and when biological connectomes have use for machine learning problems and how to properly harness connectome data for learning applications.} 

To begin answering these questions, we translate the locomotion circuit of the \textit{C. Elegans} nematode into artificial neural networks with neurons models at varying levels of biophysical realism. We use these models to solve (1) a common continuous control problem that resembles nematode locomotion in critical ways yet remains different from the details of the \textit{C. Elegans} body plan, and (2) learning problems that have little relationship to the function of the \textit{C. Elegans} locomotion circuit. For the nematode-like learning problem, we show that the \textit{C. Elegans} connectome provides a powerful inductive bias that enables improved performances over randomly connected neural networks. However, when training on learning problems with little relationship to nematode-like locomotion (e.g. image classification), the performance is dramatically hindered compared with more densely connected \textit{general} architectures.


The primary contributions of this work are as follows:

\begin{addmargin}[2.6em]{2em}

1. Results demonstrating that, while biophysical realism does indeed improve performance, it is not a necessary condition for attaining the benefits of using biological connectomes.

2. Establishing that the exact wiring diagram provides the most beneficial inductive bias, but also that retaining the architectural statistics of the biological connectome \textit{without} the precise connective patterns still proves to provide valuable priors.

3. A set of experiments demonstrating that while using biological connectomes as an architectural representation may provide a beneficial inductive bias on problems related to the function of that circuitry, it may not be beneficial \textit{in general} on problems unrelated to its function in nature. 

4. Open-source software for automatically converting connectome models into artificial neural networks that can be used for solving machine learning problems. This software includes connectomes from the \textit{C. Elegans}, which are studied in this work, and various regions of the \textit{Drosophila Melanogaster} brain, which provide exciting opportunities for future work. It also provides a general structure allowing for novel connectomes to be converted.


\end{addmargin}

We pay close attention to the training methodology (\textit{e.g. choice of optimization, population sizes that reflect natural C. Elegans populations}) and architectural design features (\textit{e.g. neuron model, synapse weight signs}) to remain as close to biology as possible while still providing useful theoretical advancements for machine learning. We believe this work provides meaningful progress toward an understanding of how connectomes may be useful and how to best harness their capabilities.

\section*{Background and Methodology}

A short introductory background on the \textit{C. Elegans} is important for understanding the results of this work, which is provided below. Here we discuss what the \textit{C. Elegans} is, how its motor circuit works to produce locomotion, and the population and evolutionary dynamics of \textit{C. Elegans} colonies.

\subsection*{The Caenorhabditis Elegans}

In the laboratory setting, \textit{C. Elegans} are one of the simplest organisms to study and thus the nematode nervous system has served as an integral model in neuroscience over the past few decades. In addition, \textit{C. Elegans} was the first organism to have its full genome sequenced and still remains as the only organism species which has its \textit{entire} neuronal wiring diagram published to date (\cite{cook2019whole}).


\textit{\textbf{The body and neural circuitry of Caenorhabditis Elegans.}} \textit{C. Elegans} is a free-living transparent nematode that thrives in temperate soil environments. Its cylindrical worm-like body spans a mere 1 \textit{mm} in length and is comprised of 959 somatic cells, 302 of which are neurons \footnote{These numbers were measured for the hermaphrodite nematode, with male nematode cell counts being generally larger (1031 somatic cells and 385 neurons).} (\cite{white1986structure}). Its exterior is unsegmented and bilaterally symmetric (\textit{i.e. worm-like}), consisting of a simple set of anatomical structures: a mouth, intestines, gonad, and connective tissue, see Fig. \ref{figure:ElegansDiag}.

Despite the incredible simplicity of these organisms, the nematode's neuronal dynamics closely resemble that of organisms with more complex nervous systems, and thus has served as a valuable model for neuroscience. Neurons in the nematode nervous system are generally separated into three categories based on their characteristic function and relationship to other neurons: sensory, motor and interneurons (\cite{cook2019whole}). In this work we focus on \textit{C. Elegans} motor neuron circuitry and apply this toward solving locomotion problems.

\begin{figure}
    \begin{center}
        \includegraphics[width=0.85\textwidth]{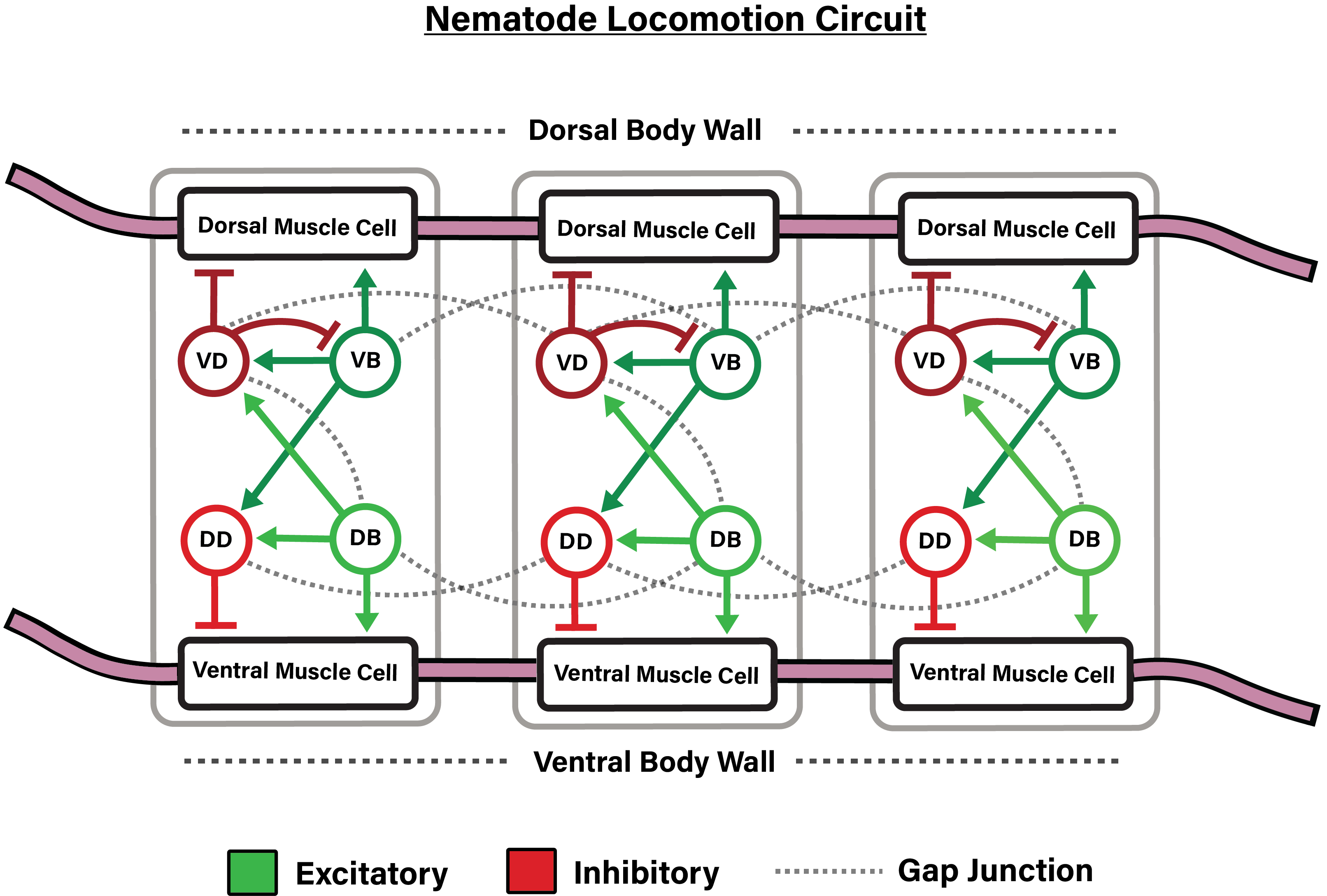}
    \end{center}
    \caption{Graphical depiction of a segment of the \textit{C. Elegans} locomotion circuit, which is characterized by the depicted microcircuits repeated down the body of the nematode. This circuit propagates waves down the nematode body which oscillate between innervating ventral and dorsal sides of the muscle body wall.}
    \label{figure:NetStructCue}
\end{figure}

\textit{\textbf{Locomotion in Caenorhabditis Elegans.}} Among the few cells in the \textit{C. Elegans} connectome, 75 are motor neurons and 95 are body wall muscle neurons. Motor neurons are stimulated by sensory and interneurons, and are the primary driver of muscle neurons activity. These neural microcircuit patterns are repeated along the body in six repeating units of ~12 motor neurons and ~12 muscle neurons, see Fig \ref{figure:NetStructCue} (\cite{haspel2011perimotor, haspel2012connectivity, zhen2015c}). These patterns produce muscle wave propagations that travel down the body and produce sinusoidal-like locomotion patterns that allow for a surprisingly wide-range of behaviors (\cite{gray2005circuit}). Within each microcircuit, forward locomotion is accomplished through the coordination of two classes of motor neurons, B and D, which are further characterized by their location along the body's radial axis, dorsal (D-) and ventral (V-). The B-type motor neuron class is excitatory, meaning it makes its outgoing neurons \textit{more likely} to fire, and acts to innervate both muscle body wall neurons along its respective radial axis and D-type neurons on the opposing radial axis (e.g. ventral to dorsal) (\cite{wen2012proprioceptive}). D-type motor neurons are inhibitory, making their outgoing connections \textit{less likely} to fire, and inhibit both the firing of their respective muscle body wall neurons and, in the case of VD neurons, the excitatory neuron along its own radial axis (\cite{wen2012proprioceptive}).

\textit{\textbf{Population dynamics of Caenorhabditis Elegans.}} Population size estimates range from 750 to 12,000 depending on the geographical location of the population, the time of year, and the availability of nutrients (\cite{frezal2015natural}). \textit{C. Elegans} populations consist of both males and hermaphrodites, however, the vast majority of individuals are \textit{hermaphrodites} (roughly 99.9\% of the species), meaning each organism has both male and female reproductive organs (\cite{chasnov2013evolutionary, frezal2015natural}). Reproduction is primarily propagated via hermaphroditic self-fertilization, where the hermaphrodite nematode produces offspring using its own reproductive material. As is discussed below, we take these principles of \textit{C. Elegans} population dynamics (e.g. hermaphrodite self-fertilization) and model them through the use of genetic algorithms.


\textit{\textbf{Genetic algorithms reflect Caenorhabditis Elegans evolutionary dynamics.}} Genetic algorithms are a biologically-inspired optimization process which frames itself around an abstraction of the process of natural selection (\cite{holland1992genetic, luke2013essentials, de2017evolutionary}). This algorithm works by (a) maintaining a population of organisms which (b) undergo a selection process based on their performance on a given task, and (c) the top performing organisms repopulate the subsequent organism pool. Selection in embodied learning problems is typically carried out through the use of a fitness function, which is a metric for determining the quality of an organism's behavior over the course of its lifetime. In a genetic algorithm, typically only the \textit{top-N} highest fitness achieving organisms survive to the next generation (referred to as \textit{elite selection}). To expand the population after the selection process, each of the remaining organisms are taken and mutated slightly to create the next generation of the population. This process reflects the evolutionary and reproductive dynamics of the \textit{C. Elegans} relatively well.

\subsection*{Solving Learning Problems with Biological Connectomes}

\textit{\textbf{Training dynamics.}} Our experimental results are collected using a genetic algorithm where replication is based on self-fertilization, or asexual reproduction, with each child in the population having only one parent (\textit{i.e. no crossover}). This is to best mirror the reproductive and evolutionary patterns of \textit{C. Elegans} populations. Additionally, we set the population size of the genetic algorithm to 750, matching \textit{C. Elegans} population ranges present in nature (\cite{barriere2005natural}).

\textit{\textbf{Connectome Representation.}} Dale's law states that in biological circuits, the sign of the weight of a synapse (excitatory or inhibitory) does not change. We wish to respect Dale's law (\cite{eccles1976electrical}) by preserving excitatory and inhibitory function of synapses in the connectome model, synaptic weights are represented on the \textit{log-scale} as follows: $w_{i,j} = s_{i,j} \text{exp}(w^{log}_{i,j})$. Genetic algorithm mutations occur on the log weights $w^{log}_{i,j}$, that appropriately bound $\text{exp}(w^{log}_{i,j})$ above zero, which, when multiplied by the original weight sign $s_{i,j}$, preserves the excitatory or inhibitory function of the synapse as is specified in the connectome model. This preservation has been shown to play significant role in retaining the functional properties of oscillatory circuits (\cite{bhattasali2022neural}).

\section*{Experiments}

\subsection*{A body design resembling Caenorhabditis Elegans locomotion}

\begin{figure}
    \begin{center}
        \includegraphics[width=0.85\textwidth]{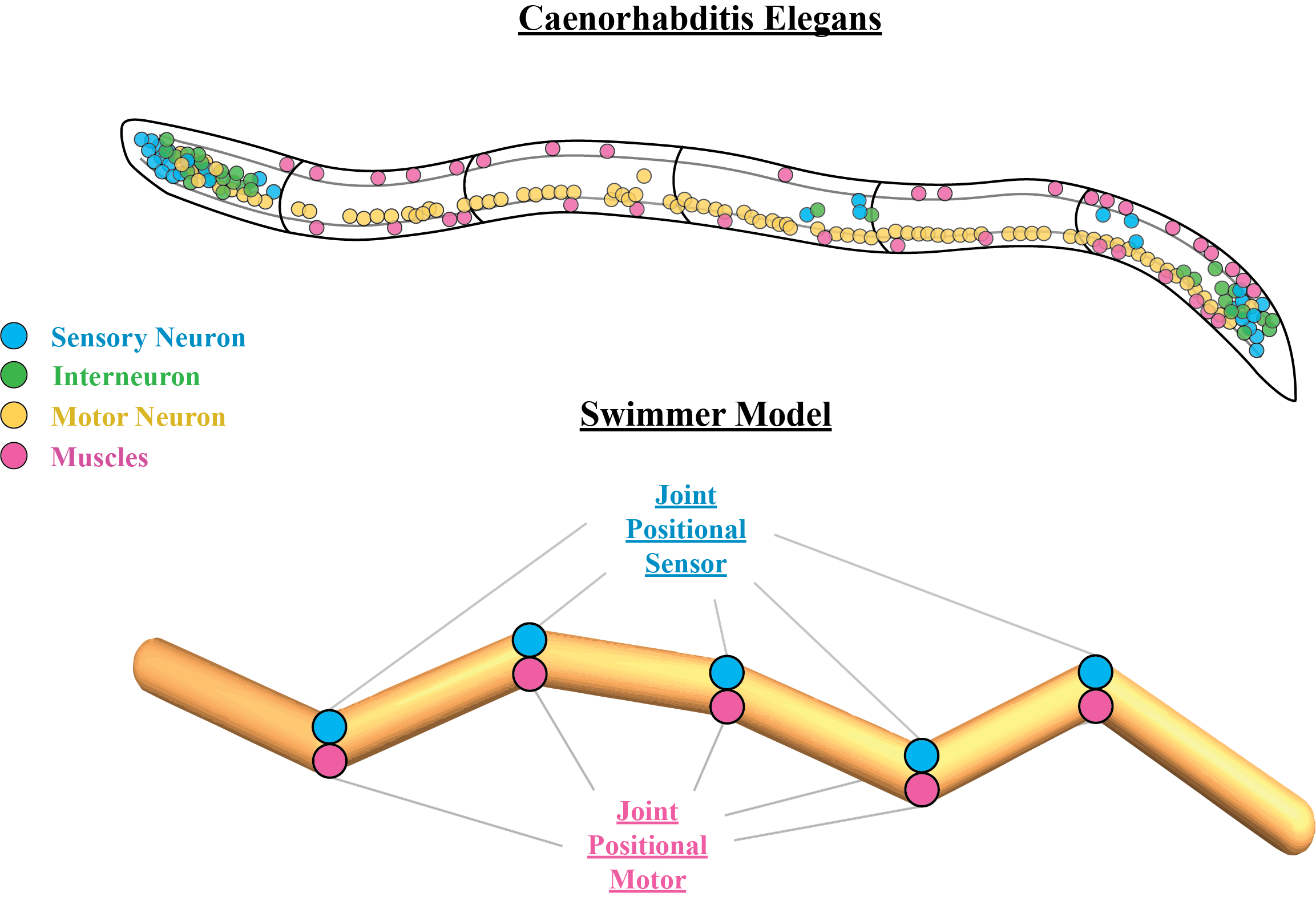}
    \end{center}
    \caption{(Top) Graphical depiction of the \textit{Caenorhabditis Elegans} nematode body and its corresponding neuronal placements categorized into sensory (\textit{blue}), motor (\textit{yellow}), muscle (\textit{pink}), and interneurons (\textit{green}). (Bottom) Depiction of the 6-jointed Swimmer body design along with its joint positional sensors and its joint positional motors highlighted blue and pink to draw relationship to \textit{C. Elegans} sensory neurons and muscle neurons respectively.}
    \label{figure:ElegansDiag}
\end{figure}

We begin our investigation of the \textit{C. Elegans} connectome on a body design, Swimmer, that is commonly used for continuous control in the Deepmind Control Suite (\cite{tassa2018deepmind}) and OpenAI Gym (\cite{brockman2016openai}). Much like in \cite{bhattasali2022neural}, we choose Swimmer instead of a more accurate neuromechanical model e.g. \cite{izquierdo2015integrated, sarma2018openworm} to demonstrate the use of connectomes on problems of interest to the machine learning community.

The Swimmer body is comprised of $6$ cylindrical links that are connected via $5$ articulated joints and looks similar to a worm, see Fig. \ref{figure:ElegansDiag}. Instead of pushing off of solid ground, the Swimmer body is encompassed in simulated fluid which it must push against to generate forward momentum and is rewarded proportional to its current velocity along the x-axis. The ideal solution in this environment is indeed very much like \textit{C. Elegans} locomotion, with the Swimmer body propagating a wave down its body starting from the head ending at the tail.

\subsection*{Varying Biophysical Realism}

Described in this experiment are three neuron models with varying degrees of biophysical realism presented in order of \textit{least} to \textit{most} biophysically accurate. We wish to determine whether we can deviate from biological neuron dynamics to still realize the benefits of the \textit{C. Elegans} connectome.


\textit{\textbf{Artificial Neuron Model.}} The artificial neuron, which is the neuron model most commonly used in machine learning applications, applies an arbitrary non-linearity to the sum of weighted inputs from pre-neuron $i$ to post-neuron $j$ via $\textbf{\textit{z}}_{j}(t) = \sigma(\textbf{\textit{w}}_{i,j}\cdot\textbf{\textit{z}}_{i}(t))$, where $\textbf{\textit{w}}_{i,j}$ represents the weight from neuron $i$ to $j$, $\textbf{\textit{z}}_{j}(t)$ represents the activity of neuron $j$ at time $t$, and $\sigma$ represents a typically smooth and continuous non-linear activation function. While the original activations used in artificial neural networks are more biological in nature (\cite{mcculloch1943logical}), these were slowly replaced over time with smoother time-invariant functions due to the use of differentiation as the medium of optimization, which tends to converge faster with smooth gradient landscapes (\cite{sharma2017activation, nwankpa2018activation}). In our experiments we use a ReLU nonlinearity, which resembles the more biologically descriptive neuron models in that it has an \textit{all-or-nothing} firing pattern, producing either a zero or some quantity \textit{x}.

\textit{\textbf{Adaptive Leaky Integrate-and-Fire Neuron Model.}} The Leaky Integrate-and-Fire (LIF) neuron is characterized via a set of equations which aim to represent the artificial neuron activation function in a more biologically realistic manner while remaining computationally efficient (\cite{lapique1907recherches, tuckwell1988introduction, abbott1999lapicque}). The LIF equations are as follows:

\begin{equation}\label{eq:LIF}
    \textbf{\textit{v}}_{j}(t+\Delta\tau) = \gamma \textbf{\textit{v}}_{j}(t) + \textbf{\textit{w}}_{i,j}(t)\textbf{\textit{s}}_{i}(t),
\vspace{1.3mm}
\end{equation}

\begin{equation}\label{eq:Spike}
    \textbf{\textit{s}}_{j}(t) = H(\textbf{\textit{v}}_{j}(t)) = \begin{dcases}
        0 & \textbf{\textit{v}}_{j}(t) < v_{th} \\
        1 & \textbf{\textit{v}}_{j}(t) \geq v_{th} \\
    \end{dcases}
.
\vspace{1.3mm}
\end{equation}

In this model, activity is integrated into the neuron membrane potential $\textbf{\textit{v}}_{j}$ and retained across time, and, once the activity exceeds a threshold value $\textbf{\textit{v}}_{j}(t) \geq v_{th}$, a binary action potential, $\textbf{\textit{s}}_{j}(t) = 1$, is propagated and the membrane potential is reset to zero. The "leaky" component of the LIF refers to the membrane potential ($\textbf{\textit{v}}_{j}$ for a given neuron $j$) being slowly decayed, or leaked, over time by a constant factor $0 < \gamma < 1$. This leaking property more accurately represents the passive ion diffusion that takes place in biological neurons, and has important consequences for learning.

While the LIF model is certainly closer to biology than the artificial neuron model, it falls short in biophysical representation capacity in several ways. The most significant disadvantage is that it cannot capture neuronal adaptation, which prevents it from representing measured spike trains with constant input current. Neuronal adaptation is the process by which, in the presence of a constant current, the time between action potentials increases over time. This process can be incorporated very simply into the LIF model, becoming the Adaptive LIF (ALIF), if we let each neuron have a local trace, $v_{adp}(t)$, which accumulates action potentials as follows: $v_{adp}(t) = \gamma_{s}v_{adp}(t) +  \boldsymbol\beta\textbf{\textit{s}}(t)$, where $\gamma_{s}$ controls the adaptation decay and $\boldsymbol\beta$ determines the affect action potentials have on the new threshold. Finally, $v_{adp}(t)$ is incorporated into the action potential firing threshold which becomes $\textbf{\textit{v}}_{j}(t) > v_{th}$ becomes $\textbf{\textit{v}}_{j}(t) > v_{th} + v_{adp}(t)$. 

However, like the LIF model, the ALIF model still struggles to capture fundamental behaviors present in biological neurons (\cite{izhikevich2004model}) and thus falls short in providing thorough biophysical accuracy.


\textit{\textbf{Izhikevich Neuron Model.}} Many accurate biophysical neurons models, such as the Hodgkins-Huxley, are computationally prohibitive and can only simulate several neurons at real-time speed, preventing their use in computationally intensive applications. To get around this problem, while still maintaining the important properties of more complicated biophysical models, the Izhikevich neuron model (\cite{izhikevich2003simple}) effectively balances realism and tractability. This is accomplished via the following set of differential equations:

\begin{equation}\label{eq:VoltageIzhik}
    v_{i}(t+\Delta\tau) = v_{i}(t) + 0.04v_{i}^{2} + 5v_{i} + 140 - u_{i} + \sum_{j}W_{i,j}(t)s_{j}(t).
\vspace{1.3mm}
\end{equation}

\begin{equation}\label{eq:CurrentIzhik}
    u_{i}(t+\Delta\tau) = u_{i}(t) + a(bv_{i} - u_{i}).
\vspace{1.3mm}
\end{equation}

\begin{wrapfigure}{l}{0.6\textwidth}
  \begin{center}
    \includegraphics[width=0.6\textwidth]{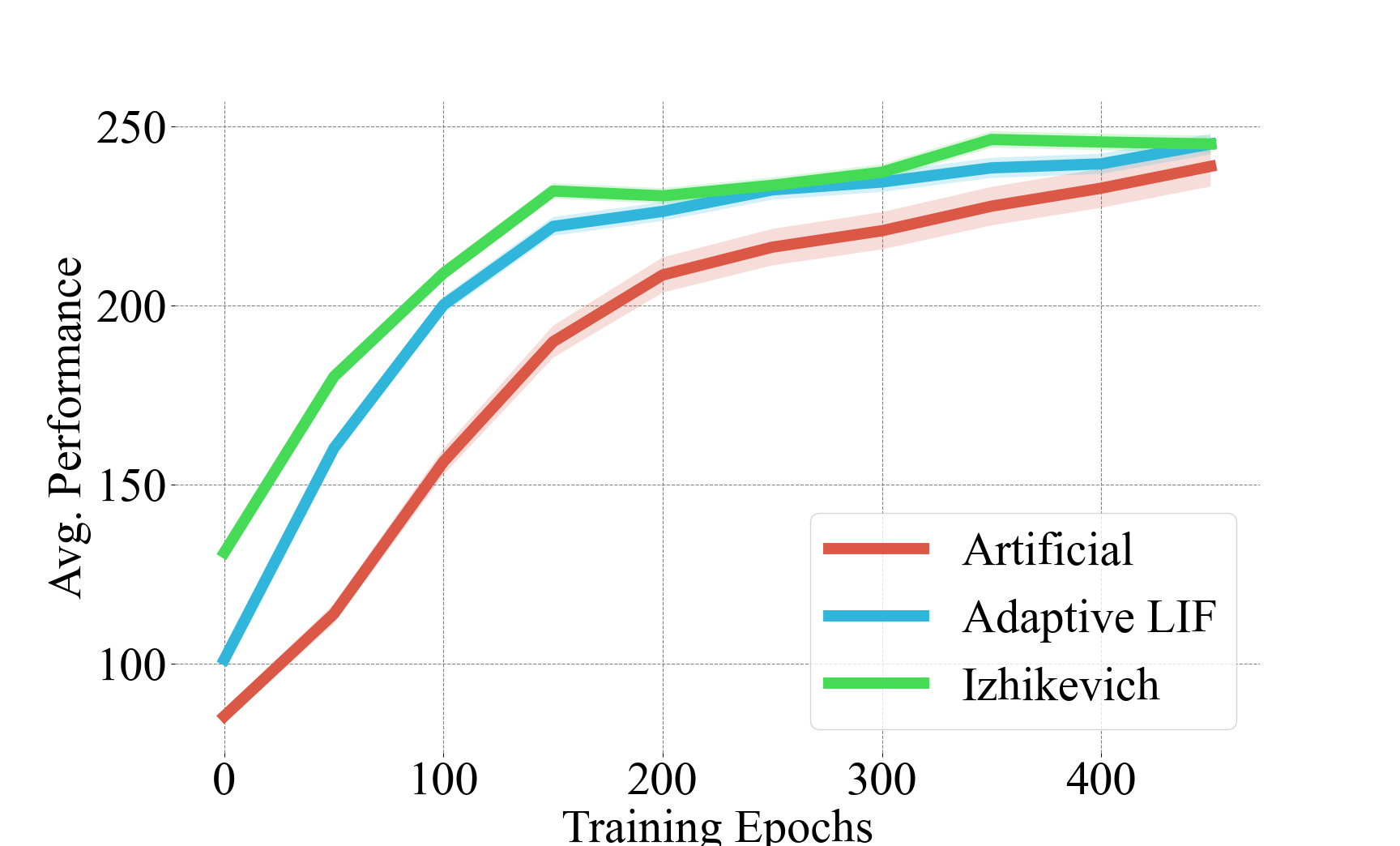}
  \end{center}    
  \caption{Performance across }
  \label{figure:Biophys}
\end{wrapfigure}

In these equations, a spike is fired ($s_j(t) = 1$) when $v_{i}(t) \geq 30$ mV, upon which $v \leftarrow c$ and $u \leftarrow u+d$. While seemingly non-biologically descriptive at first, the Izhikevich neuron model is actually the product of rigorously fitting the parameters of a quadratic integrate-and-fire equation to a wide range of \textit{in-vitro} recordings of different cortical neuron dynamics. This allows the model to be both highly biophysically accurate and computationally efficient. A selection of the values of the four variables \textit{a, b, c, and d} determine the spiking and bursting dynamics of the neuron. Additional training details for all three neuron models are provided in the Appendix.

\textit{\textbf{Results.}} The premise of the following experiment is simple: We will compare the performance of the \textit{C. Elegans} locomotion circuit with varying degrees of realism trained on the Swimmer locomotion problem. The weights of these circuits are trained using a genetic algorithm which resembles \textit{C. Elegans} evolutionary dynamics as is described both in \textit{Methods} and in-depth in the \textit{Appendix}. The results of these experiments are shown in Fig. \ref{figure:Biophys}. 

Similar to the artificial circuits used in \cite{bhattasali2022neural}, we do not observe much additional performance growth relative to the network prior with additional training, especially with the more biologically realistic neuron models. We can see that the three networks converge to similar final values, with the more biophysically accurate neuron models performing slightly better than the artificial neuron model. The primary difference between these models is in their \textit{initial average performance}, with the Izhikevich neuron model beginning at $\sim130$, the ALIF around $\sim100$ and the Artificial neuron model around $\sim50$. The most important finding of these graphs however, is that biophysical realism is not necessary to realize the potential of biological circuits. 



\subsection*{Architectural statistics conserve benefits of inductive priors}

While a comprehensive map of the \textit{C. Elegans} neural connections is available in its entirety, this organism is an exception -- it is the only organism with a complete connectomic mapping. Nonetheless, while no other organism has a complete mapping, many organisms have \textit{approximate connectome statistics} available (\textit{e.g. number of neurons, number of inhibitory/excitatory synapses}). To motivate the use of connectomes, it would also be necessary to motivate the use of \textbf{connectome statistics} since outside of the \textit{C. Elegans} this constitutes the majority of our knowledge of the connectomes of other organisms.
\begin{wrapfigure}{r}{0.5\textwidth}
  \begin{center}
    \includegraphics[width=0.5\textwidth]{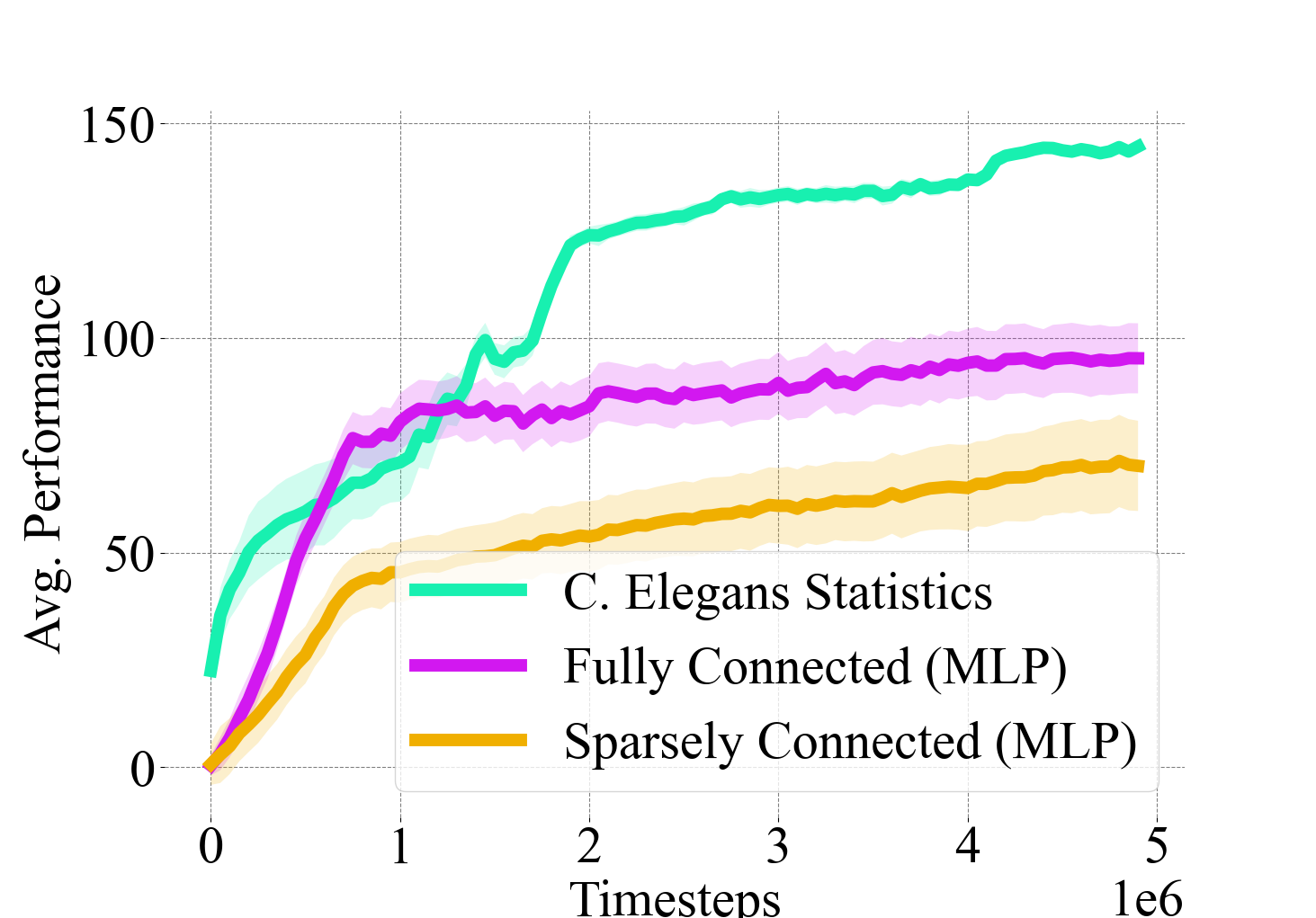}
  \end{center}    
  \caption{Performance across three network architectures. The first of which has the connectivity statistics of the C. ELegans locomotion circuit. The second is a fully connected MLP. The final curve represents a sparsely connected MLP with randomly sampled sparsity.}
  \label{figure:StatisticCurves}
\end{wrapfigure}

Toward this end, we design an experiment where the \textbf{essential statistics} of the \textit{C. Elegans} locomotion circuit are used to construct a sparse Multi-Layer Perceptron (MLP). We then compare the performance of the \textit{C. Elegans} approximate MLP with a fully connected MLP and a sparsely connected MLP with random architectural statistics. The performance is compared on the Swimmer task in order to determine how much architectural bias the \textit{C. Elegans} circuit retains.

The \textit{C. Elegans} statistics network is a sparse artificial neural network which has the same number of neurons and synapses as the original \textit{C. Elegans} connectome, where the excitatory and inhibitory neuron ratios are preserved but the precise circuitry is not. The fully connected neural network is a two-hidden-layer neural network with the same number of neurons as the \textit{C. Elegans} network, however the neurons have a fully connected structure between layers (i.e. an MLP). The final network is also a two-hidden-layer MLP, however the layer between hidden neurons has a sparsely connected topology, where the sparsity and excitatory-inhibitory ratios are randomly sampled from a uniform distribution.

Fig. \ref{figure:StatisticCurves} shows the performance across the three compared network architectures. Both the fully connected and sparsely connected MLPs begin with a performance around $0$, but the approximate \textit{C. Elegans} network begins with a performance around $22$, retaining a fraction of the locomotion circuit inductive bias. The \textit{C. Elegans} network obtains a final performance of $146.4\pm 3.5$ compared with the fully connected MLP around $96.2 \pm 13.6$ and the sparsely connected MLP around $63.7 \pm 17.2$. It is clear that the precise circuitry of the \textit{C. Elegans} connectome remains superior. However, these results indicate that, \textbf{while the precise connectome may not be known, approximate architectural statistics can still provide useful information for solving a problem.} Nonetheless, more investigation needs to be done around why connectome statistics still provide an inductive bias despite the network microcircuits being degraded.

\subsection*{The limitations of architectural priors}

We have thus far demonstrated that the \textit{C. Elegans} connectome model may provide useful behavioral priors on \textit{a particular task} that is closely related to the purpose of that connectome. In this case, we showed that the \textit{C. Elegans} connectome provides a powerful inductive prior on a swimmer locomotion task that requires oscillatory wave propagation motion -- exactly what the \textit{C. Elegans} locomotion circuit is designed to do. However, we wish to investigate to what extent these priors prove useful on tasks that do not resemble their biological purpose. How well would the \textit{C. Elegans} connectome perform on a legged locomotion problem or on an image learning problem?

To determine this, we examine the performance of the \textit{C. Elegans} connectome on a locomotion problem from the same benchmark suite, as well as a classification problem that is completely outside the domain of the \textit{C. Elegans} architecture design. For the locomotion problem, we evaluate the performance of the connectome on the Half Cheetah task (\cite{tassa2018deepmind}), which consists of a 2D planar robot with a total of 7 links, 2 legs and a torso, where each leg has a total of 3 actuated joints. The objective is to run forward with the highest velocity possible. For the image learning problem we evaluate the performance of the connectome on MNIST (\cite{lecun1998mnist}), a 28x28 dimensional handwritten digit classification problem with 10 digit classes.

\begin{figure}[h]
    \begin{center}
        \includegraphics[width=0.98\textwidth]{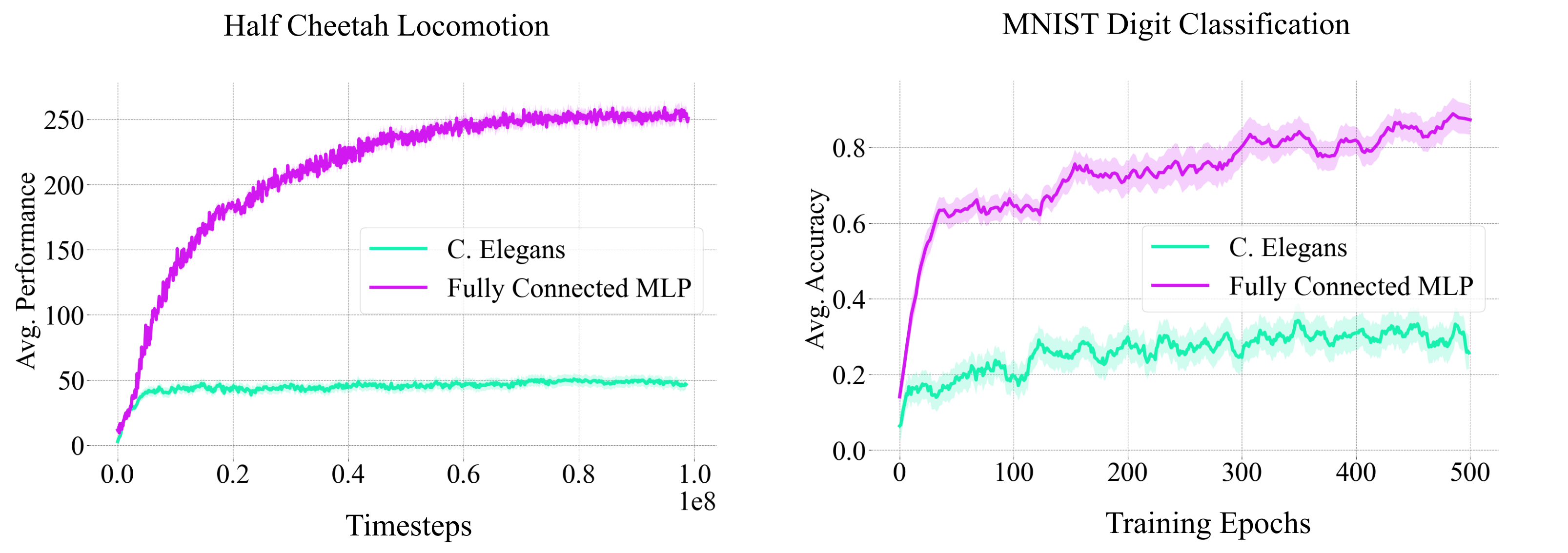}
    \end{center}
    \caption{Training performance curves on (a) a reinforcement learning problem, the Half Cheetah locomotion benchmark (\cite{tassa2018deepmind}), and (b) an image learning problem, MNIST  (\cite{lecun1998mnist}). It can be seen that the C. Elegans network structure hinders performance on both tasks and converges early in the training process, whereas the MLP performs relatively well on both tasks.}
    \label{figure:PerfVariety}
\end{figure}

In Fig. \ref{figure:PerfVariety} we can see that on both the Half Cheetah and MNIST problems, the C. Elegans connectome not only does not outperform a standard MLP like in previous problems, but also performs dramatically worse than it. There is a characteristic deterioration in performance growth after a small amount of training time. This lack of performance growth may be a product of the \textit{C. Elegans} dependence on structure priors at birth versus on intra-lifetime learning like in higher order mammals, thus developing an architecture that biases toward \textit{innate knowledge} over \textit{learning capability}. While this innate bias is helpful for the Swimmer locomotion problem, as these results demonstrate, this innate bias is in fact a limitation when using the connectome on problems that do not resemble nematode locomotion. Stated more clearly, the architecture constrains the set of possible behaviors, in turn biasing the network to the subset of behaviors that the architecture was designed to perform (e.g. \textit{oscillations}). These findings demonstrate that connectomes must be applied in the right context to realize the benefits of their architectural priors.



\section*{Discussion and Future Work}

In this work we discerned where and when biological connectomes can prove advantageous for machine learning. We showed that for problems resembling the function of the connectome (\textit{e.g. oscillatory locomotion for the C. Elegans)}, there exists a strong and beneficial inductive bias. However, for problems not resembling the function of the connectome, its structure constrains the model's representation, and degrades training performance. We showed that increasing biological realism improves the priors of the connectome, but that realism is not necessary for extracting the benefits of the network structure. Finally, we demonstrated that architectural statistics conserve some benefits of the inductive priors of the architecture.

One of the primary limitations of using biological connectomes is that it requires a degree of understanding in systems neuroscience to enable an effective implementation. To alleviate this difficulty, we have developed open-source software tools to automatically convert connectome models into artificial neural networks that can be readily used for solving optimization problems\footnote{The link will be attached here upon acceptance.}. This includes a process of finding a good set of \textit{initial shared weights} for the excitatory and inhibitory connections, as well as tuning the neuron parameters for arbitrary learning problems. We hope this software will help further research at the intersection of connectomics and machine learning, particularly for exploring the connectome of the \textit{Drosophila melanogaster}, which includes a visual processing system (\cite{takemura2015connectome}) that may lead to new and powerful architectural discoveries for visual learning problems.

We believe that connectomes may provide structure that promotes stable bounds for models of synaptic plasticity, which have seen a recent resurgence for use in deep learning applications (\cite{miconi2020backpropamine, najarro2020meta, schmidgall2021spikepropamine, schmidgall2022learning}). As this work has demonstrated, the connectome seems to provide a strong inductive bias on both the initial behavior and the set of possible behaviors that the network can produce. The use of structured connectomes could prevent highly dynamic weights from deteriorating behavior. This differs from a more general architecture where the behavior would change much more dramatically with small changes in weights (\cite{schmidgall2022stable}). These perturbations become more dramatic when weights are changing simultaneously and independently. 

We believe that the incredible sparsity of biological connectomes (\textit{3.2\% for the C. Elegans}, \cite{cook2019whole}) will significantly decrease the amount of energy necessary to deploy deep learning models, perhaps through the use of neuromorphic hardware (\cite{young2019evolving, zhu2020comprehensive, schuman2022opportunities}). These connectomes may lead to more robust and resilient neural systems that sidestep many of the adversarial drawbacks of highly general network structures (\cite{guo2018sparse, schuman2020resilience}). Finally, we believe that the use of biological connectomes in machine learning may lead to profound advancements in both our understanding of the role of neural architecture and in discovering new neural architectures. We believe that our work is a step toward bridging the gap between our understanding of systems neuroscience and artificial intelligence.

\section*{Reproduciblity}

To ensure reproducibility we have included an exhaustive list of hyperparameters for both the GA, ES, and neuron models in every experiment. A detailed explanation of the training process is also described in the Appendix. Finally, a software toolkit for using connectomes as neural architecture is released along with the paper.

\bibliography{iclr2021_conference}
\bibliographystyle{iclr2021_conference}

\appendix
\section{Appendix}

\subsection{Hardware Specifications}

\textbf{Training Device:} 2021 Apple MacBook Pro

\textbf{Operating System:} macOS Monteray v12.3

\textbf{Processor}: Apple M1 Max with 10-core CPU (8 Performance and 2 Efficiency)

\textbf{Installed Physical Memory (RAM):} 64 GB of Unified Memory (LPDDR5)

\subsection{Training details}

Inputs were mapped into the locomotion circuit motor neurons (or the hidden neurons for MLPs) via a pre-trained non-linear encoder for each of the three networks (Artificial, ALIF, Izikevich) and for the three compared connectivity statistics networks. Outputs were mapped from the muscle body-wall neuron spiking activity to motor actions (or classification labels) via a pre-trained non-linear encoder. Additionally, for all networks, the initial weights and neuron model constants were pretrained, like in previous work (\cite{bhattasali2022neural}), and these initial weights were individually fine-tuned during the GA training. Pretraining was accomplished using an Evolutionary Strategies optimizer (\cite{salimans2017evolution}) for 1000 epochs, with details below.

Neural network, genetic algorithm, and evolutionary strategies code was implemented using the Numpy Python library (\cite{harris2020array}).

\subsection{Hyperparameters}

\subsubsection{Genetic Algorithm} 

\hspace{4mm}\textit{\textbf{Varying Biophysical Realism}}

\hspace{6mm}\textbf{Population Size:} $750$

\hspace{6mm}\textbf{Elite Selection Size:} $8$

\hspace{6mm}\textbf{Mutation Rate (Encoder/Decoder):} 0.01

\hspace{6mm}\textbf{Mutation Rate (Connectome Weights):} 0.01

\hspace{6mm}\textbf{Mutation Rate Decay (Encoder/Decoder):} 0.997

\hspace{6mm}\textbf{Mutation Rate Decay (Connectome Weights):} 0.997

\hspace{4mm}\textit{\textbf{Architectural Statistics Conserve Benefits of Indudctive Priors}}

\hspace{6mm}\textbf{Population Size:} $750$

\hspace{6mm}\textbf{Elite Selection Size:} $8$

\hspace{6mm}\textbf{Mutation Rate (Encoder/Decoder):} 0.01

\hspace{6mm}\textbf{Mutation Rate (Connectome Weights):} 0.02

\hspace{6mm}\textbf{Mutation Rate Decay (Encoder/Decoder):} 0.997

\hspace{6mm}\textbf{Mutation Rate Decay (Connectome Weights):} 0.997

\hspace{4mm}\textit{\textbf{Limitations of Architectural Priors}}

\hspace{6mm}\textbf{Population Size:} $750$

\hspace{6mm}\textbf{Elite Selection Size:} $8$

\hspace{6mm}\textbf{Mutation Rate (Encoder/Decoder):} 0.02

\hspace{6mm}\textbf{Mutation Rate (Connectome Weights):} 0.02

\hspace{6mm}\textbf{Mutation Rate Decay (Encoder/Decoder):} 0.998

\hspace{6mm}\textbf{Mutation Rate Decay (Connectome Weights):} 0.998

\subsubsection{Evolutionary Algorithm} 

\hspace{6mm}\textbf{Population Size:} 200

\hspace{6mm}\textbf{Mutation Rate (Encoder/Decoder):} 0.1

\hspace{6mm}\textbf{Mutation Rate (Shared, Connectome Weights):} 0.1

\hspace{6mm}\textbf{Learning Rate (Encoder/Decoder):} 0.1

\hspace{6mm}\textbf{Learning Rate (Shared, Connectome Weights):} 0.1

\hspace{6mm}\textbf{Mutation Rate Decay (Encoder/Decoder):} 0.999

\hspace{6mm}\textbf{Mutation Rate Decay (Shared, Connectome Weights):} 0.999

\hspace{6mm}\textbf{Learning Rate Decay (Encoder/Decoder):} 0.999

\hspace{6mm}\textbf{Learning Rate Decay (Shared, Connectome Weights):} 0.999

\subsubsection{Neuron Models} 

\hspace{4mm}\textit{\textbf{Adaptive Leaky Integrate-and-Fire Neuron Model}}

\hspace{6mm}\textbf{Voltage Decay:} $e^{-1/20}$

\hspace{6mm}\textbf{Adaptation Time Constant:} $0.5$

\hspace{6mm}\textbf{Adaptive Threshold Decay:} $e^{-1/10}$

\hspace{6mm}\textbf{Connectome Weight Initialization Sample (Excitatory):} $w_{i,j} = e^{-1.0}$

\hspace{6mm}\textbf{Connectome Weight Initialization Sample (Inhibitory):} $w_{i,j} = e^{-1.5}$

\hspace{6mm}\textbf{Encoder/Decoder Weight Initialization Sample:} $w_{i,j} \sim $U$(-0.3, 0.3)$

\hspace{4mm}\textit{\textbf{Izhikevich Neuron Model}}

\hspace{6mm}\textbf{Initial Parameter \textit{a}:} $0.02$

\hspace{6mm}\textbf{Initial Parameter \textit{b}:} $0.25$

\hspace{6mm}\textbf{Initial Parameter \textit{c}:} $-58$

\hspace{6mm}\textbf{Initial Parameter \textit{d}:} $0$

\hspace{6mm}\textbf{Spike Firing Threshold:} $30$

\hspace{6mm}\textbf{Synaptic Delay:} $1$ ms

\hspace{6mm}\textbf{Current Scalar} $20.0$

\hspace{6mm}\textbf{Voltage Constant Scalar} $0.2$

\hspace{6mm}\textbf{Adaptive Threshold Decay:} $e^{-1/10}$

\hspace{6mm}\textbf{Connectome Weight Initialization Sample (Excitatory):} $w_{i,j} = e^{2.0}$

\hspace{6mm}\textbf{Connectome Weight Initialization Sample (Inhibitory):} $w_{i,j} = e^{1.0}$

\hspace{6mm}\textbf{Encoder/Decoder Weight Initialization Sample:} $w_{i,j} \sim $U$(-0.3, 0.3)$

\hspace{4mm}\textit{\textbf{Artificial Neuron Model}}

\hspace{6mm}\textbf{Connectome Weight Initialization Sample (Excitatory):} $w_{i,j} = e^{-2.5}$

\hspace{6mm}\textbf{Connectome Weight Initialization Sample (Inhibitory):} $w_{i,j} = e^{-3.0}$

\hspace{6mm}\textbf{Encoder/Decoder Weight Initialization Sample:} $w_{i,j} \sim $U$(-0.3, 0.3)$

\end{document}

%% file: math_commands.tex
\usepackage{amsmath,amsfonts,bm}

\def\eqref#1{equation~\ref{#1}}

\def\1{\bm{1}}

\DeclareMathAlphabet{\mathsfit}{\encodingdefault}{\sfdefault}{m}{sl}
\SetMathAlphabet{\mathsfit}{bold}{\encodingdefault}{\sfdefault}{bx}{n}

